\documentclass[10pt,twocolumn,letterpaper]{article}

\usepackage{cvpr}              % To produce the CAMERA-READY version

\usepackage{pifont}
\usepackage{soul}
\usepackage{colortbl}
\usepackage{comment}
\usepackage{wrapfig, lipsum, booktabs}
\usepackage{makecell}
\usepackage{enumitem}
\usepackage{amsmath}
\usepackage{nicematrix}
\usepackage{adjustbox}
\usepackage{multirow}
\usepackage{bm}

\definecolor{ForestGreen}{RGB}{0, 179, 45}

\mathchardef\mhyphen="2D

\newcommand{\vhtext}{\bm{h}_{\rm text}}
\newcommand{\vhbox}{\bm{h}_{\rm box}}
\newenvironment{my_enumerate}%
  {\begin{enumerate}\vspace{-5pt}
    \setlength{\itemsep}{5pt}%
    \setlength{\parskip}{0pt}}%
  {\vspace{-5pt}\end{enumerate}}
\definecolor{cvprblue}{rgb}{0.21,0.49,0.74}
\usepackage[pagebackref,breaklinks,colorlinks,citecolor=cvprblue]{hyperref}

\def\paperID{11972} % *** Enter the Paper ID here
\def\confName{CVPR}
\def\confYear{2024}

\title{Boximator: Generating Rich and Controllable Motions for Video Synthesis}

\author{Jiawei Wang$^{*}$ ~~~Yuchen Zhang$^{*}$ ~~~Jiaxin Zou \\
~~~Yan Zeng  ~~~Guoqiang Wei ~~~Liping Yuan ~~~Hang Li \\ \\
ByteDance Research \\
{\tt\small $^{*}$ Equal Contribution} \\
{\tt\small \{wangjiawei.424, zhangyuchen.zyc, zoujiaxin.zjx, }\\
{\tt\small zengyan.yanne, weiguoqiang.9, yuanliping.0o0 lihang.lh\}@bytedance.com} \\
\href{https://boximator.github.io}{\texttt{https://boximator.github.io}}
}

\begin{document}
\maketitle
\begin{figure*}[ht]
\centering
\includegraphics[width=\textwidth]{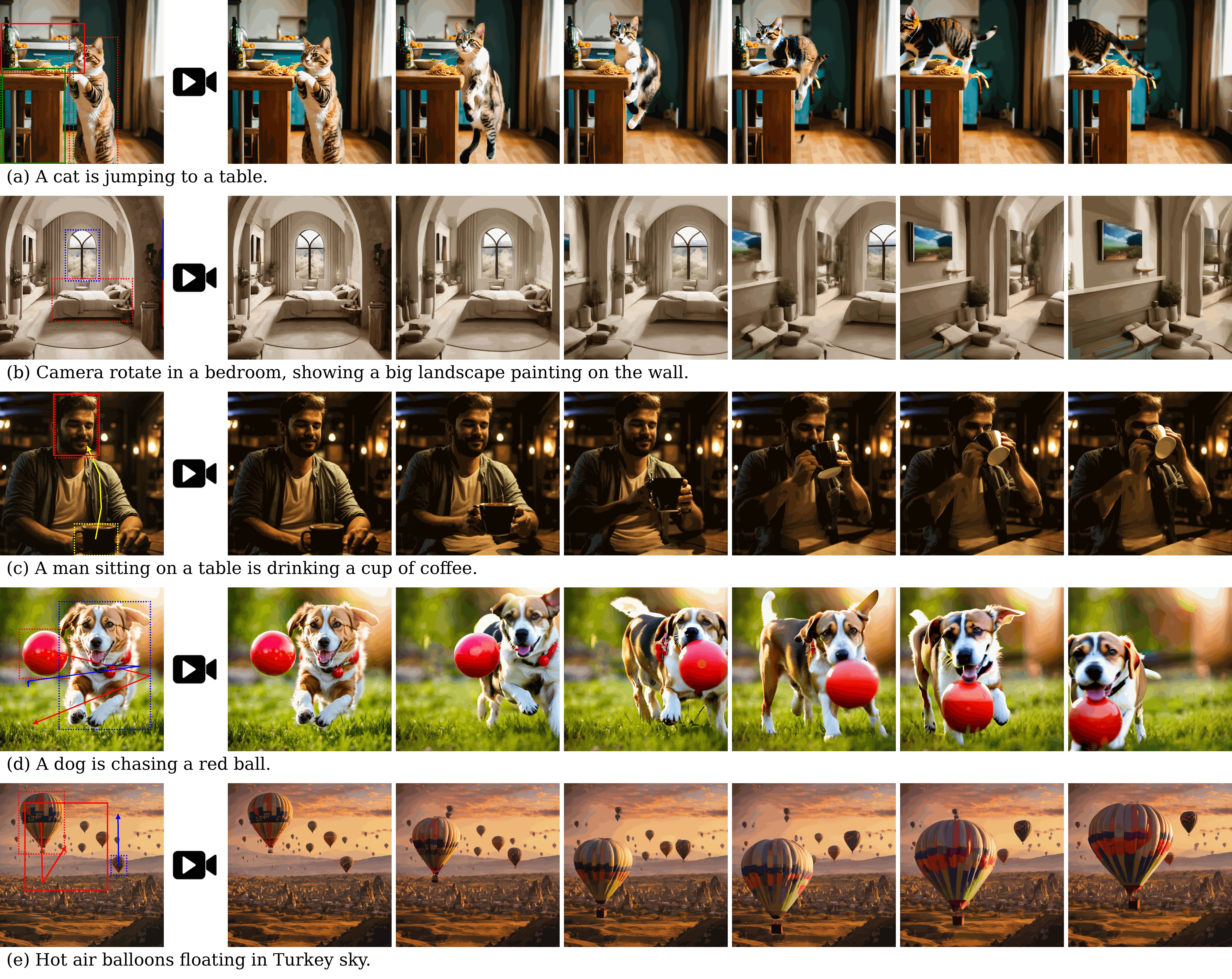}
\caption{Motion control with Boximator: (a) use hard boxes to control the ending shape and position of a jumping cat; (b) force camera rotating to the left by pushing bed and window to the right; (c) control how a person raises a cup of coffee; (d) control the motion path of a dog and a ball; (e) Use motion path and hard boxes to control the trajectory and proximity of two balloons. In all figures, dotted boxes represent first frame constraints; solid-line boxes represent last frame constraints; Arrowed lines represent motion paths. Example videos are initially generated as 256x256x16, then enhanced to 768x768x16 via a pretrained super-resolution model.}
\label{fig:opening}
\vspace{-10pt}
\end{figure*}

\begin{abstract}
Generating rich and controllable motion is a pivotal challenge in video synthesis. We propose \textbf{Boximator}, a new approach for fine-grained motion control. Boximator introduces two constraint types: \textbf{hard box} and \textbf{soft box}. Users select objects in the conditional frame using hard boxes and then use either type of boxes to roughly or rigorously define the object's position, shape, or motion path in future frames. Boximator functions as a plug-in for existing video diffusion models. Its training process preserves the base model's knowledge by freezing the original weights and training only the control module. To address training challenges, we introduce a novel \textbf{self-tracking} technique that greatly simplifies the learning of box-object correlations. Empirically, Boximator achieves state-of-the-art video quality (FVD) scores, improving on two base models, and further enhanced after incorporating box constraints. Its robust motion controllability is validated by drastic increases in the bounding box alignment metric. Human evaluation also shows that users favor Boximator generation results over the base model.
\end{abstract}

\section{Introduction}

Video synthesis has recently experienced remarkable advancements~\cite{ho2022video,ho2022imagen,singer2022make,girdhar2023emu,ge2023preserve,kondratyuk2023videopoet}. These models typically utilize either a text prompt or a key frame to generate videos. Recent research focuses on enhancing the controllability by incorporating frame-level constraints, such as sketches, depth maps~\cite{wang2023videocomposer,guo2023sparsectrl}, human poses~\cite{xu2023magicanimate,feng2023dreamoving,wang2023disco}, trajectories~\cite{yin2023dragnuwa,wang2023motionctrl}, and conditional images~\cite{qing2023hierarchical,zeng2023make,chen2023seine}.

In this work, we introduce a novel approach utilizing box-shaped constraints as a universal mechanism for fine-grained motion control. Our method introduces two types of constraints: \emph{hard box}, which precisely delineates an object's bounding box, and \emph{soft box}, defining a broader region within which the object must reside. The soft box can be as tight as the object's exact bounding box, or as loose as the frame boundary. We control multiple objects across frames by associating unique object IDs with these boxes. Our proposed method, named Boximator (combining ``box'' and ``animator''), offers several benefits:

\begin{my_enumerate}

\item Boximator serves as a flexible motion control tool. It manages the motion of both foreground and background objects, as well as modifies the pose of larger objects (e.g., human) by adjusting smaller components. Refer to Figure~\ref{fig:opening} for illustrations.

\item In scenarios where generation is conditioned on an image, as seen in image-to-video and many state-of-the-art text-to-video methods~\cite{zeng2023make, girdhar2023emu}, users can easily select objects by drawing hard boxes around them. This visually-grounded approach is more straightforward compared to the language-grounded controls~\cite{huang2023fine,ma2023trailblazer}, which require verbal descriptions for all objects.

\item For frames lacking user-defined boxes, Boximator allows approximate motion path control via algorithm-generated soft boxes. These soft boxes can be constructed based on a pair of user-specified boxes, or based on a hard box combined with a user-specified motion path. See Figure~\ref{fig:opening}(c)-(e) for examples of user-specified motion paths.

\end{my_enumerate}

Boximator functions as a plug-in for existing video diffusion models. We encode every box constraint by four coordinates, an object ID, and a hard/soft flag. During training, we freeze the base model's text encoder and U-Net, feeding the box encoding through a new type of self-attention layer. This design is inspired by GLIGEN~\cite{li2023gligen}, where bounding boxes are combined with object description texts to achieve region control for image synthesis. However, Boximator aims to control object motions without relying on textual grounding, thus requires the learning of box-object correlation purely from visual inputs.

Empirically, we find that it is hard for the model to learn this visual correlation through standard optimization. To mitigate this challenge, we introduce an novel training technique termed \emph{self-tracking}. This technique trains the model to generate colored bounding boxes as a part of the video. It simplifies the challenge into two easier tasks: (1) producing a bounding box for each object with the right color, and (2) aligning these boxes with the Boximator constraints in every frame. We observe that video diffusion models can quickly master these tasks. After that, we train the model to stop generating visible bounding boxes. Although these boxes are no longer visually present, their internal representation persists, enabling the model to continue aligning with Boximator constraints.

We developed an automatic data annotation pipeline to generate 1.1M highly dynamic video clips with 2.2M annotated objects from the WebVid-10M dataset~\cite{bain2021frozen}. We utilized this dataset to train our Boximator model on two base models: the PixelDance model~\cite{zeng2023make} and the open sourced ModelScope model~\cite{wang2023modelscope}. Extensive experiments show that Boximator retains the original video quality of these models while offering robust motion control in diverse real-world scenarios. On the MSR-VTT dataset, Boximator improves upon the base models in FVD score. With box constraints added, video quality significantly improved further (PixelDance: 237 $\rightarrow$ 174, ModelScope: 239 $\rightarrow$ 216), and the object detector's average precision (AP) score, measuring box-object alignment, saw a remarkable increase (1.9-3.7x higher on MSR-VTT and 4.4-8.9x higher on ActivityNet), highlighting effective motion control. User study also favored our model's video quality and motion control over the base model by large margins (+18\% for video quality, +74\% for motion control). Furthermore, ablation studies confirm the necessity of introducing soft boxes and training with self-tracking for achieving these results.

\section{Related Work}

Video diffusion models are natural extensions of image diffusion models. They extend the U-Net architecture from image models by adding temporal layers~\cite{ho2022imagen,singer2022make}. A widely adopted method for improving computational efficiency is to denoie in the latent space~\cite{he2022latent,zhou2022magicvideo}. Text-to-video (T2V) diffusion models are often the foundation for various forms of conditional generation \cite{ge2023preserve,blattmann2023stable,wang2023modelscope}. Recent advancements suggest a two-step approach to T2V: initially creating an image based on text, followed by producing a video that considers both the text and the pre-generated image. This approach allows the video model to concentrate on dynamic aspects by using a static image as a reference, leading to improved video quality \cite{zeng2023make,girdhar2023emu,wang2024magicvideo}. The reference image provides a natural grounding source for motion control.

There is a surge in research focused on enhancing the controllability of T2V and I2V models. VideoComposer~\cite{wang2023videocomposer} enables conditions such as sketches, depth maps, and motion vectors. In producing dance videos, human poses extracted from reference videos are commonly used~\cite{xu2023magicanimate,feng2023dreamoving,wang2023disco}. For more precise motion control, users can plot object or camera trajectories~\cite{yin2023dragnuwa,wang2023motionctrl}. However, these methods did not provide a precise way to define objects, making it challenging to select and control a larger or composite object from image. Moreover, trajectory does not capture the object's shape and size, crucial for depicting pose or proximity changes like arm spreading or approaching movements.

There are two concurrent research studying the use of bounding boxes for motion control, but it should be noted that their work differs from ours in key aspects. TrailBlazer~\cite{ma2023trailblazer} is a training-free method that leverages attention map edits to direct the model in generating a specific object within a designated area. The object must be described in the text prompt. FACTOR~\cite{huang2023fine} modified a transformer-based generation model, Phenaki~\cite{villegas2022phenaki}, by adding a box control module. Like TrailBlazer, FACTOR requires a text description for each box, thus does not support visual grounding. Neither of the above methods supports soft box constraints, nor do they study the associated challenges in training.

\section{Background: Video Diffusion Model}

Boximator is built on top of video diffusion models~\cite{ho2022video} using the 3D U-Net architecture~\cite{ronneberger2015u}. These models iteratively predict the noise vector in noisy video inputs, gradually transforming pure Gaussian noise into high-quality video frames. The U-Net, denoted by $\epsilon_\theta$, processes a noisy input $z$ (either in pixel space or latent space), along with a timestamp $t$ and various conditions $c$, and predicts the noise in $z$. Optimization is achieved through a noise prediction loss:
\[
\mathcal{L}_\theta = \mathbb{E}_{z_0,c,\epsilon,t} [\|\epsilon - \epsilon_\theta(z_t,t,c)\|_2^2],
\]
where $z_0$ represents the ground truth video, $\epsilon$ is a Gaussian noise vector, and $z_t$ is a noisily transformed version of $z_0$:
\[
z_t = \sqrt{\bar\alpha_t}z_0 + \sqrt{1-\bar\alpha_t}\epsilon.
\]
Here, $\bar\alpha_t$ denotes a predefined constant sequence.

The 3D U-Net is structured with alternating convolution blocks and attention blocks. Each block comprises two components: a spatial component, handling individual video frames as separate images, and a temporal component, enabling information exchange across frames. Within every attention block, the spatial component typically includes a self-attention layer, followed by a cross-attention layer, the latter used for conditioning the generation on a text prompt. 

\section{Boximator: Box-guided Motion Control}

\subsection{Model Architecture}\label{sec:architecture}

\begin{figure}
\centering 
\includegraphics[width=\columnwidth]{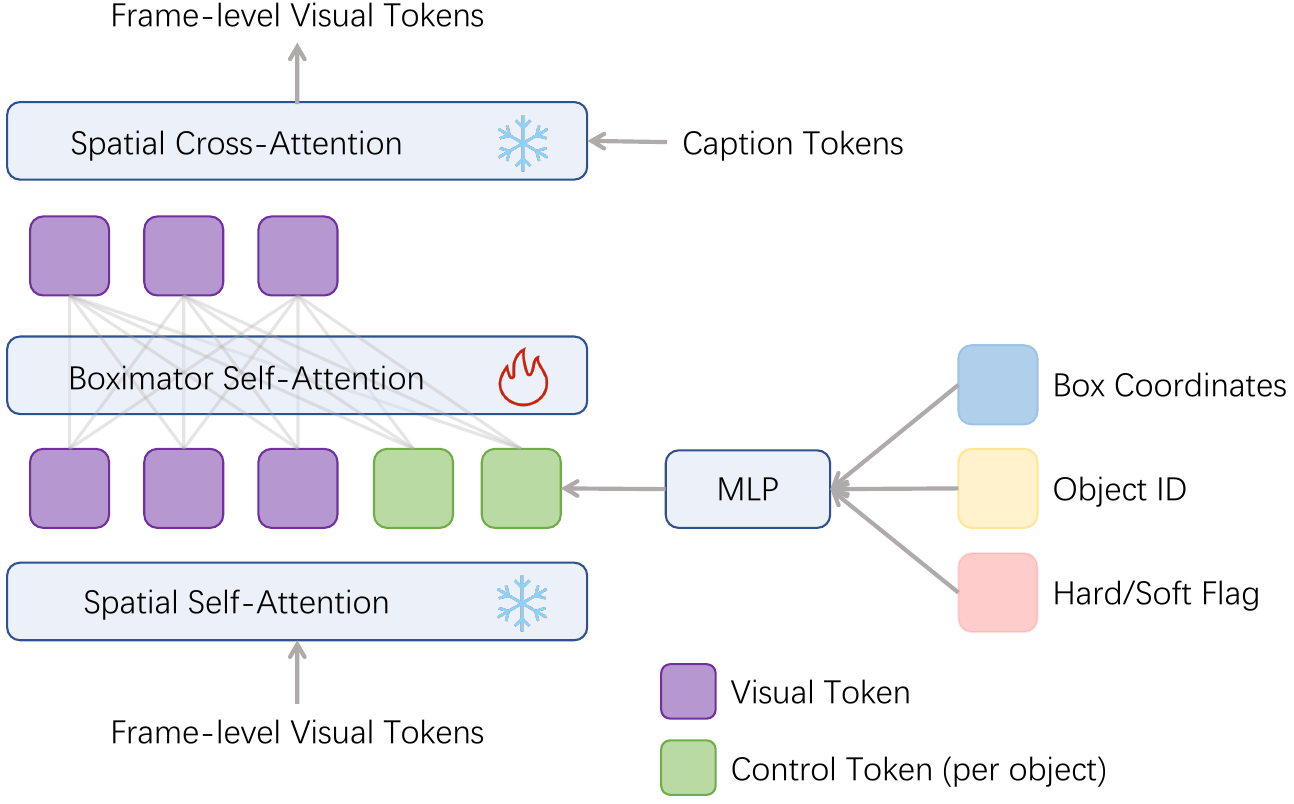}
\caption{Overview of the control module: adding a new self-attention layer to every spatial attention block, between the spatial self-attention and the spatial cross attention. During training, all the original model parameters are frozen.}
\label{fig:model}
\vspace{-10pt}
\end{figure}

Our objective is to endow existing video diffusion models with motion control capabilities. Given that foundation models are pre-trained on extensive collections of web-scale images and videos, it's crucial to preserve their acquired knowledge. To achieve this, we freeze the original model parameters and solely focus on training the newly incorporated motion control module.

The architecture of our model is illustrated in Figure~\ref{fig:model}. In every spatial attention block of video diffusion models, there are two stacked attention layers: a spatial self-attention layer and a spatial cross-attention layer. We augment this stack by adding a new self-attention layer. Specifically, if $\bm{v}$ denotes the visual tokens of a frame, and $\vhtext$ and $\vhbox$ represent the embeddings of the text prompt and the box constraints, respectively, then the modified spatial attention block is described as follows:
\begin{align*}
    \bm{v} &= \bm{v} + {\rm SelfAttn}(\bm{v})\\
    \bm{v} &= \bm{v} + {\rm TS}({\rm SelfAttn}([\bm{v}, \vhbox]))\\
    \bm{v} &= \bm{v} + {\rm CrossAttn}(\bm{v}, \vhtext)
\end{align*}
where TS(·) is a token selection operation that exclusively considers visual tokens. The box embeddings $\vhbox$ is a sequence of control tokens. Each token represents a box and is defined by:
\[
\bm{t}_b = {\rm MLP}({\rm Fourier}([b_{\rm loc}, b_{\rm id}, b_{\rm flag}])).
\]
Here, $b_{\rm loc}$ is a 4-dimensional vector encapsulating the top-left and bottom-right coordinates of the box, normalized to the range [0,1]. The object ID, used to link boxes across frames, is represented by $b_{\rm id}$, which in our experiments is expressed in RGB space. Each object thus corresponds to a unique ``color'' for its boxes, making $b_{\rm id}$ a 3-dimensional vector normalized to [0,1]. The $b_{\rm flag}$ is a boolean indicator: it is set to 1 for hard boxes and 0 otherwise. These three inputs are concatenated and processed via a Fourier embedding~\cite{mildenhall2021nerf} followed by a multi-layer perceptron (MLP). Note $\vhbox$ contains a fixed number of control tokens (indicated by $N$). When the actual number of boxes is smaller than $N$, we use a learnable $\bm{t}_{\rm null}$ to pad the empty slots.

\subsection{Data Pipeline}\label{sec:data-pipeline}

In the absence of a publicly available video dataset with object tracking annotations, we curated our training set from the WebVid-10M dataset~\cite{bain2021frozen}. Through empirical analysis, we find that a vast majority of WebVid videos do not exhibit substantial object or camera movements. Consequently, sampling from this collection would be inefficient for training our motion control module. To address this issue, we curated a more dynamic subset from WebVid. This involved evaluating every clip in the dataset, comparing their start and end frames, and retaining only those clips where the two frames are sufficiently different. This filtration yielded a total of 1.1M video clips.

For every clip in our refined dataset, we took the first frame to generate an image description using a visual language model~\cite{liu2023visual}. Then we extract noun chunks from these descriptions. These chunks, encompassing terms like "young man" or "white shirt," served as object prompts. We then feed these prompts to a pre-trained grounding model~\cite{liu2023grounding} and object tracker~\cite{cheng2023tracking} to generate bounding boxes and populate them across all frames of the video. This approach successfully yielded bounding boxes for a total of 2.4M objects.

\begin{figure}
\centering 
\includegraphics[width=0.7\columnwidth]{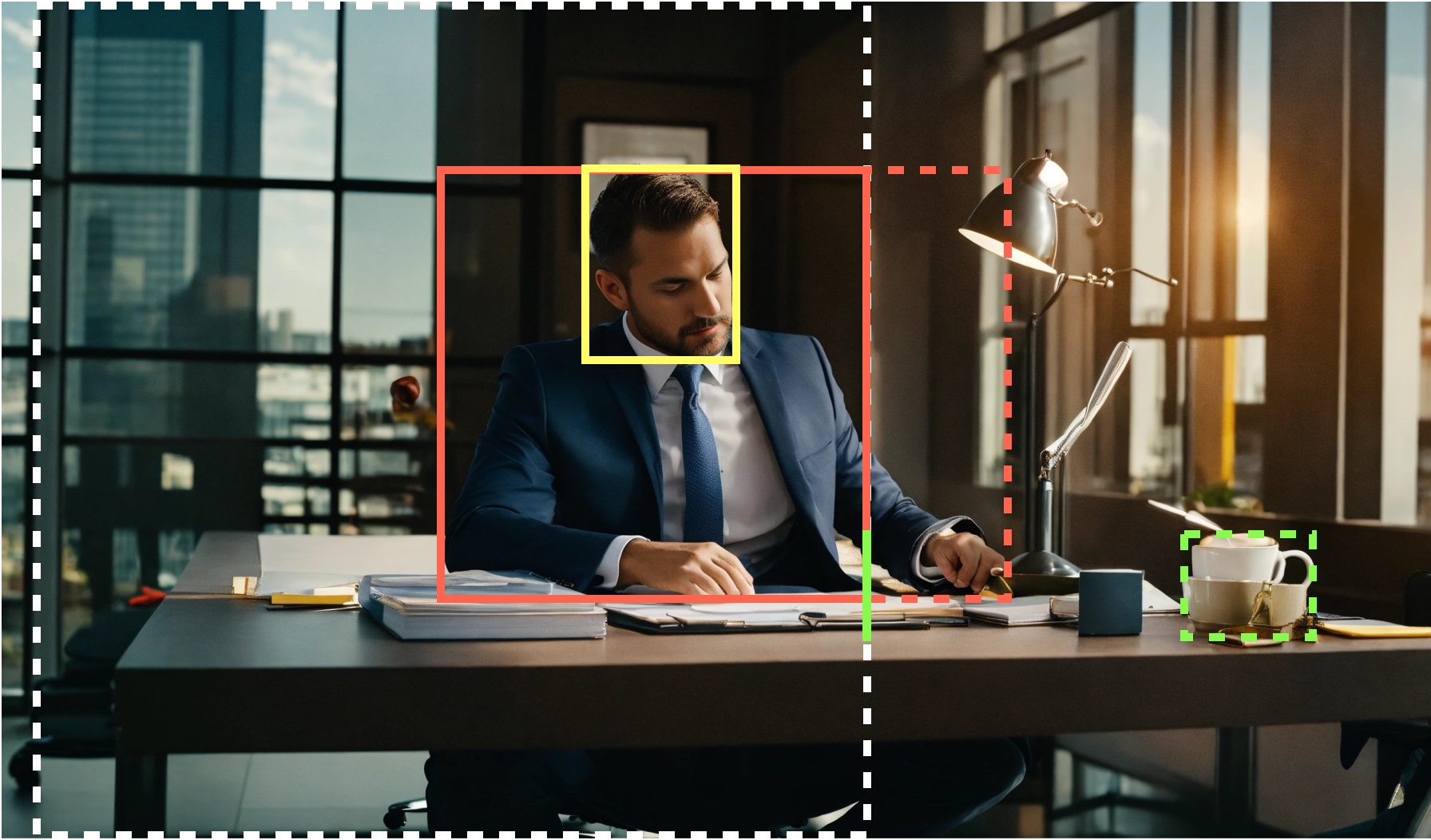}
\caption{Training data: all bounding boxes are projected to the cropped region (white dashed box).}
\label{fig:projection}
\end{figure}

During training, we take a random crop of the video, conforming to the specified target aspect ratio, and subsequently project all bounding boxes onto this cropped region (Figure~\ref{fig:projection}). If a bounding box entirely fall outside the cropped area, then we project it as line segments along the border of the crop. This allows users to control object movements both into and out of the frame by drawing line segments on the frame's border (See Figure~\ref{fig:case-study}(d) for an example).

\begin{figure}
\centering 
\includegraphics[width=\columnwidth]{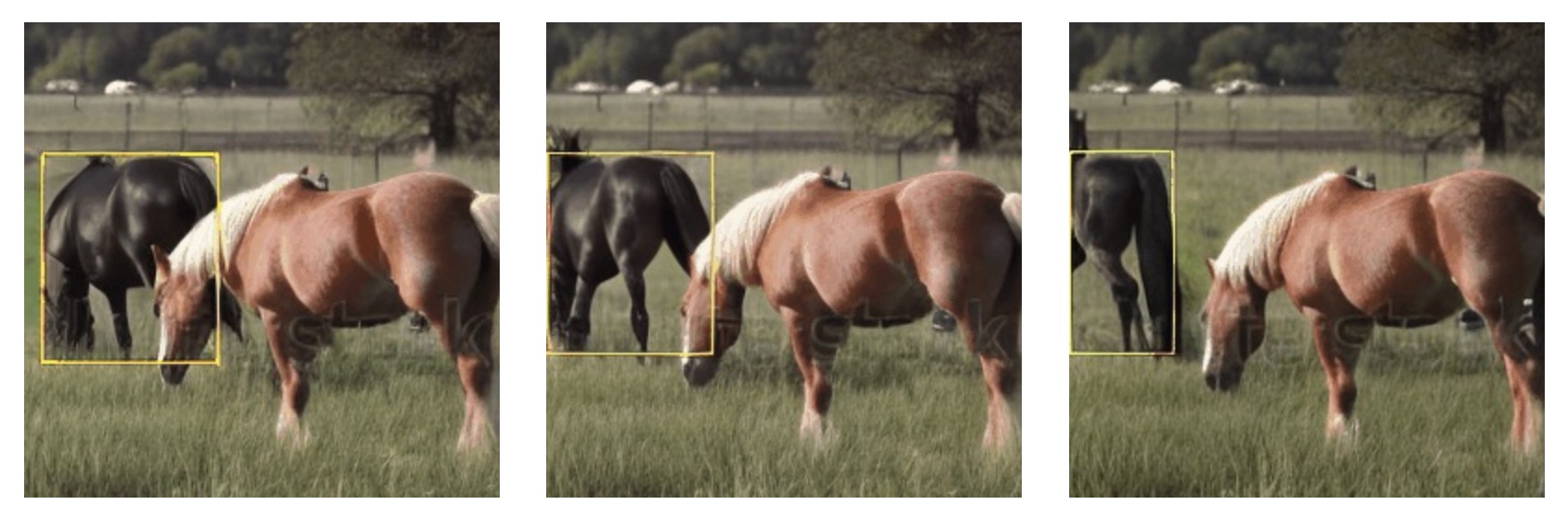}
\caption{Self-tracking: train the model to track every constrained object. This figure shows 3 frames where the black horse and the yellow box surrounding it are generated together.}
\label{fig:selfgrounding}
\vspace{-10pt}
\end{figure}

\subsection{Self-Tracking}~\label{sec:self-tracking}

A significant challenge in video motion control lies in associating box coordinates with objects and maintaining temporal consistency across frames, namely making sure that the same group of boxes always control the same object. In practice, this proves to be challenging, as diffusion models often struggle to effectively link discrete control signals, like coordinates and IDs, with visual elements. This difficulty is exacerbated when the video contains multiple overlapping boxes. As Section~\ref{sec:ablation} shows, with traditional loss optimization, the model failed to align to most box constraints after 110K steps of training.

We propose self-tracking as a simple technique to mitigate this challenge. We train our model to generate colored bounding boxes for each constrained object in every frame, with colors specified in the object's control token (Figure~\ref{fig:selfgrounding}). In other words, we train the model to perform generation and object tracking at the same time. This approach simplifies the problem into two easier tasks: (1) generating a bounding box for each object with the right color and (2) aligning these boxes with the Boximator constraints in every frame. Previous research in image synthesis~\cite{sheynin2023emu} shows that diffusion models can generate bounding boxes. We further discover that diffusion models can maintain temporal consistency, ensuring that boxes of the same color consistently track the same object across frames. With this capability, task (2) becomes straightforward. For hard box constraints, the model only needs to put boxes at the specified coordinates, while for soft box constraints, it needs to put them within a specified region. Intuitively, self-tracked bounding boxes act as an intermediary representation. The model follows Boximator constraints to guide the generation of these boxes, which in turn guide the generation of objects.

Upon completing the self-tracking training phase, we proceed to further train the model using the same dataset, but excluding bounding boxes from the target frames. Remarkably, the model quickly learn to cease generating visible bounding boxes, but its box alignment ability persists. This indicates that the self-tracking phase assists the model to develop an appropriate internal representation.

\subsection{Multi-Stage Training Procedure}\label{sec:training-procedure}

We employ a multi-stage training procedure. Initially, in Stage 1, the model is trained using all the provided ground truth bounding boxes as hard box constraints. Since hard box controls are easier to learn than the soft ones, this stage serves as a preliminary phase, establishing the model's initial understanding of coordinates and IDs. Subsequently, in Stage 2, we substitute 80\% of these hard boxes with soft boxes. The soft boxes are generated by randomly and independently expanding the hard ones in four directions: left, right, up, and down. The expansion margin for each direction is determined by a ${\rm Beta}(1,8)$ distribution, so that the average expansion is 1/9 of the frame's width or height, while the maximum expansion can extend up to the frame's boundary. Both Stage 1 and Stage 2 use the self-tracking technique outlined in Section~\ref{sec:self-tracking}. Finally, in Stage 3, we continue the Stage 2 training but without self-tracking.

\begin{figure}
\centering 
\includegraphics[width=\columnwidth]{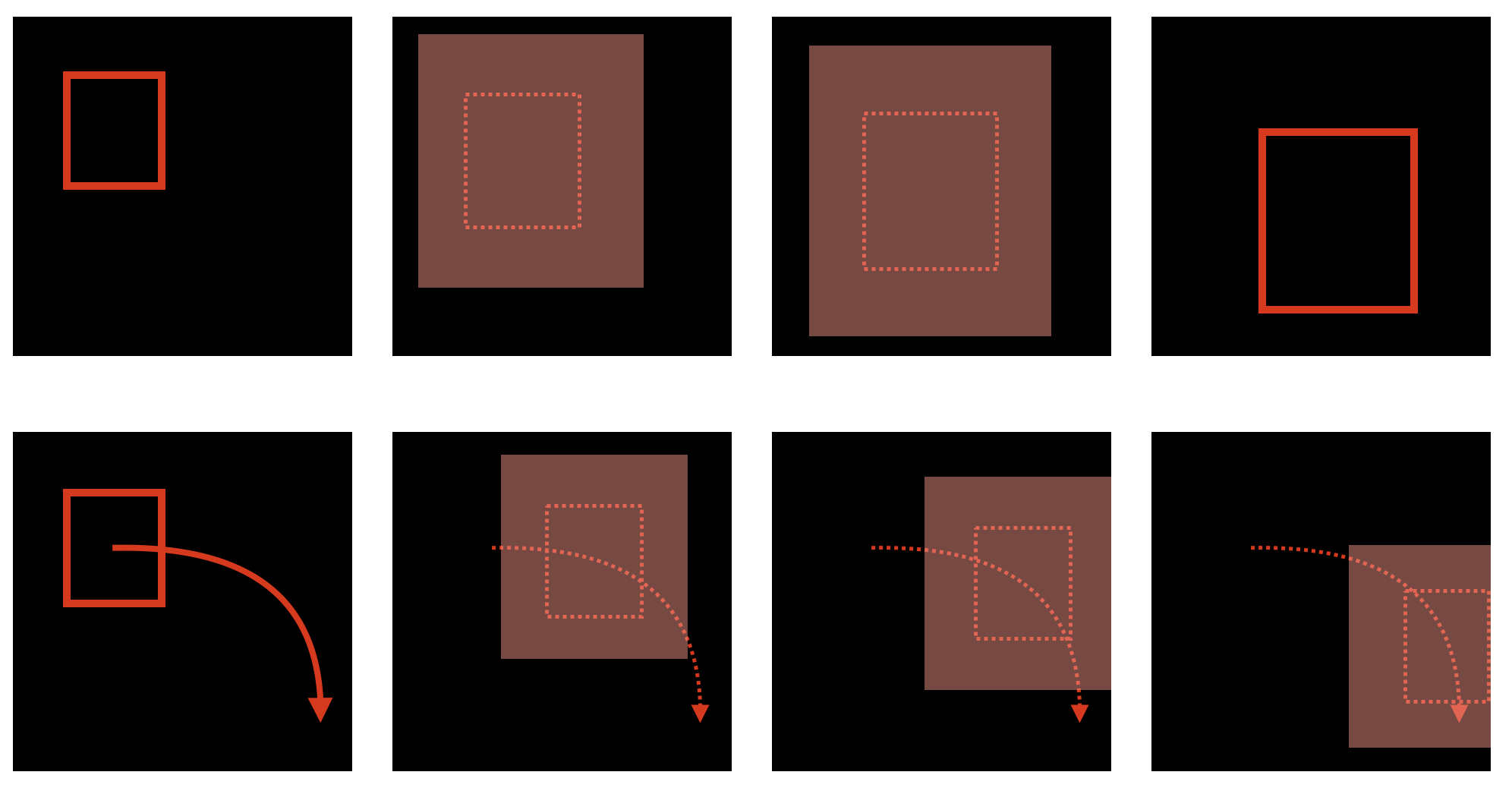}
\caption{Soft boxes in inference. We interpolate soft boxes and relax them based on a pair of user-specified boxes (upper row), or a user-specified box combined with a motion path (lower row).}
\label{fig:interpolation}
\vspace{-10pt}
\end{figure}

\subsection{Inference}\label{sec:inference}

During the inference stage, only a select few frames (such as the first and last) contain user-defined boxes. To achieve robust control, we insert soft boxes to the other frames. This is done by first applying linear interpolation of user-defined boxes to those empty frames, and then relaxing the interpolated boxes by expanding the box regions (as described in Section~\ref{sec:training-procedure}) and marking them as ``soft box''. This approach ensures that the object roughly follows the intended trajectory, while simultaneously offering the model sufficient flexibility to introduce variations. In cases where a user draws a hard box in a frame and defines a motion path for it, we let the box to slide along the path to construct interpolated boxes for each subsequent frame, then relax them to form soft box constraints. Figure~\ref{fig:interpolation} presents a visual illustration for the construction of soft boxes in both cases. 

\section{Experiments}
\subsection{Experiment Settings}\label{sec:experiment-settings}

\paragraph{Base models} We train Boximator on two base models: PixelDance~\cite{zeng2023make} and ModelScope~\cite{wang2023modelscope}. Our experiments use text prompts, box constraints, and optionally the first video frame as input conditions. PixelDance can directly use the first frame as an input condition. ModelScope doesn't support direct image input, but we can still condition it on an image by replacing the first frame of the noisy latents $z$ with the ground-truth frame's latents at each denoising step. In both cases, we freeze the original model weights and only train the control module. See Appendix~\ref{sec:appendix-implementation-details} for more training and inference details.

\paragraph{Datasets} We test our models using the MSR-VTT~\cite{xu2016msr}, ActivityNet~\cite{caba2015activitynet} and UCF-101\footnote{The UCF-101 dataset details and results are discussed in Appendix~\ref{sec:appendix-ucf}}~\cite{soomro2012ucf101} datasets. MSR-VTT test set consists of 2,990 samples with 20 prompts
per example. For text constraint, following~\cite{huang2023fine, zeng2023make}, we randomly select one prompt per sample to generate one video. For box constraint, MSR-VTT does not include bounding box annotations, so we automatically create reference (ground-truth) bounding boxes. First, we identify noun chunks from the text prompt. Then, we use Grounding DINO~\cite{liu2023grounding} to get bounding boxes on the first frame and DEVA~\cite{cheng2023tracking} to extend these boxes to subsequent frames. 

Considering that the automatic annotations on MSR-VTT may be noisy, to increase the credibility of our results, we manually annotated a portion of the ActivityNet validation set. Specifically, we chose 796 video clips that include noticeable object motion. The bounding boxes in the first frame have already been annotated by the ActivityNet Entities dataset~\cite{zhou2019grounded}, and we manually extended the bounding box annotations to all 16 frames. 

\paragraph{Evaluation metrics} We measure video quality using Fréchet Video Distance (FVD)~\cite{unterthiner2018towards} and measure text alignment using CLIP similarity score (CLIPSIM)~\cite{wu2021godiva}. We compute the FVD metrics using the randomly selected 16 frames of each ground truth video with an FPS of 4. For evaluating motion control, we use the average precision (AP) metric. We generate videos with ground-truth boxes on the first and last frames as constraints. After creating a video, we detect bounding boxes with the aforementioned DINO+DEVA detection system. If an object is consistently tracked across all frames, we compare its detected bounding box with the ground truth boxes on the first/last frame. AP is calculated following the MS COCO protocol~\cite{lin2014microsoft}. When the first frame is a given condition, we only compare boxes on the last frame. We also report mean average precision (mAP), calculated as the average AP over 10 Intersection over Union (IoU) thresholds, from 0.5 to 0.95.

\subsection{Quantitative Evaluation}

\begin{table*}
\centering
\small
\begin{tabular}{l l|c c c}
\toprule
\textbf{Models} & Extra Input & FVD($\downarrow$) & CLIPSIM($\uparrow$) & mAP/AP$_{50}$/AP$_{75}$($\uparrow$) \\ \hline
MagicVideo~\cite{zhou2022magicvideo} & - & 1290 & - & - \\
LVDM~\cite{he2022latent} & - & 742 & 0.2381 & -\\ 
ModelScope~\cite{wang2023modelscope} & - & 550 & 0.2930 & -\\
Show-1~\cite{zhang2023show} & - & 538 & 0.3072 & -\\ 
PixelDance~\cite{zeng2023make} & - & 381 & {\bf 0.3125} & -\\
Phenaki~\cite{villegas2022phenaki} & - & 384 & 0.2870 & -\\ 
FACTOR-traj~\cite{huang2023fine} & Box & 317 & 0.2787 & 0.290$^*$/-/- \\ \hline
\hline
\multirow{4}{*}{PixelDance + Boximator} & - & {\bf 237} & 0.3039 & 0.094/0.193/0.076 \\
                  & Box & 174 & 0.2947 & 0.349/0.479/0.359 \\ 
                  & F0 & 113 & 0.2890 & 0.194/0.330/0.177 \\
                  & F0 + Box & 102 & 0.2874 & 0.365/0.521/0.384 \\ \hline
\multirow{4}{*}{ModelScope + Boximator} & - & 239 & 0.3013 & 0.096/0.195/0.084 \\
                  & Box & 216 & 0.2948 & 0.312/0.470/0.309 \\
                  & F0  & 142 & 0.2865 & 0.141/0.260/0.126 \\
                  & F0 + Box &  132  & 0.2852 & 0.300/0.456/0.299 \\ \bottomrule
\end{tabular}
\caption{Zero-shot results on MSR-VTT. \textbf{F0} means given the first frame as condition. \textbf{Box} means box constraints. The results show that Boximator retains or improves the video quality (FVD) of the base models. In all cases, adding box constraints (Box) significantly improves the average precision (AP) score of bounding box alignment.}
\label{table:msr-vtt-results}
\end{table*}

\paragraph{Video Quality}

Table~\ref{table:msr-vtt-results} compares our models with recent video synthesis models on the MSR-VTT dataset. In text-to-video synthesis, our Boximator model outperforms the base models, achieving competitive FVD scores of 237 and 239 with PixelDance and ModelScope, respectively. This improvement, despite using frozen base model weights, is probably due to the control module's training on motion data, enhancing dynamic scene handling.

The results in Table~\ref{table:msr-vtt-results} indicates that the FVD score improves when extra conditions are added to the input. Specifically, the introduction of box constraints (Box) enhances video quality (PixelDance: 237 $\rightarrow$ 174; ModelScope: 239 $\rightarrow$ 216). We hypothesize this improvement is due to box constraints providing a more realistic layout for video generation. However, when the generation is based on the first frame (F0), the impact of box constraints on FVD is reduced (PixelDance: 113 $\rightarrow$ 102; ModelScope: 142 $\rightarrow$ 132). This might be because the layout is already set by F0.

Our models achieve CLIPSIM scores that are on par with state-of-the-art systems. We noticed a slight drop in CLIPSIM scores when additional conditions (like F0 or Box) are introduced. This occurs because the base model is optimized for aligning video with the text alone, whereas our model handles multiple types of alignment at the same time. A similar observation was reported in the FACTOR paper~\cite{huang2023fine}.

\paragraph{Motion Control Precision}

Table~\ref{table:msr-vtt-results} also presents the results for motion control precision. In every case, adding box constraints (Box) significantly improves the average precision (AP) scores. This indicates that the model effectively understands and applies the box constraints. The FACTOR paper~\cite{huang2023fine} reported the mAP score on MSR-VTT too. Although our results aren't directly comparable to theirs due to differences in object annotations, we've included their number (marked with $^*$) in Table~\ref{table:msr-vtt-results} for reference.

Table~\ref{table:anet-results} presents the results on ActivityNet. We intentionally chose test videos from ActivityNet that feature significant object movements. As a result, the disparity in AP scores before and after adding box constraints is wider compared to MSR-VTT.
The mAP scores with box constraints are 4.4-8.9x higher than that without box on ActivityNet, in contrast to 1.9-3.7x higher on MSR-VTT.

It's important to note that the AP scores in our experiments are not equal to success rate in motion control. To calculate AP, we compare the reference object boxes with those generated by the video synthesis model and detected by the DINO+DEVA system. This detector isn't flawless; it might miss objects, detect irrelevant ones, or fail to track an object consistently across all frames. These potential errors in detection can impact the final AP score. Therefore, it's more insightful to focus on the difference in AP scores between methods, rather than the absolute values.

\begin{table}
\centering
\begin{tabular}{l l | c}
\toprule
\small
\textbf{Base Models} & Extra Input & mAP/AP$_{50}$/AP$_{75}$($\uparrow$) \\\hline
\multirow{4}{*}{PixelDance} & - & 0.050/0.103/0.041\\
                  & Box & 0.445/0.638/0.459 \\
                  & F0 & 0.079/0.165/0.072 \\
                  & F0 + Box & 0.394/0.607/0.409 \\ \hline
\multirow{4}{*}{ModelScope} & - & 0.054/0.118/0.040 \\
                  & Box & 0.361/0.563/0.372 \\
                  & F0  & 0.069/0.128/0.068 \\
                  & F0 + Box & 0.304/0.522/0.291 \\ \bottomrule
\end{tabular}
\caption{Box alignment results on ActivityNet. In all cases, adding box constraints significantly improves the AP score.}
\label{table:anet-results}
\end{table}

\begin{table}
\centering
\small
\resizebox{\columnwidth}{!}{
\begin{tabular}{l|c c c}
\toprule
\textbf{Criteria} & Boximator wins & Draw & Base model wins \\\hline
 Video Quality  & 35.2\% & 48.0\% & 16.8\% \\
 Motion Control & 76.0\% & 21.8\% & 2.2\% \\ \bottomrule
\end{tabular}
}
\caption{Human side-by-side blind comparison on 100 samples.}
\label{table:human-eval}
\end{table}

\begin{table}
\centering
\resizebox{\columnwidth}{!}{
\begin{tabular}{l|c c}
\toprule
\small
\textbf{Methods} & mAP (Box) & mAP (F0+Box) \\\hline
 & \multicolumn{2}{c}{MSR-VTT} \\ 
PixelDance + Boximator & 0.349 & 0.365\\
\quad w/o self-tracking & 0.118 & 0.187\\
\quad w/o soft boxes & 0.235 & 0.274\\
\quad w/o freezing weights & 0.354 & 0.343 \\ \hline \hline
& \multicolumn{2}{c}{ActivityNet} \\
PixelDance + Boximator & 0.445 & 0.394 \\
\quad w/o self-tracking & 0.083 & 0.085 \\
\quad w/o soft boxes & 0.248 & 0.220 \\
\quad w/o freezing weights & 0.404 & 0.331 \\ \bottomrule
\end{tabular}
}
\caption{Ablation study: removing self-tracking and soft boxes both result in significant drop in the box alignment metric. Training all model weights doesn't give extra benefits.}
\label{table:ablation}
\end{table}

\begin{figure*}[ht]
\centering
\includegraphics[width=0.93\textwidth]{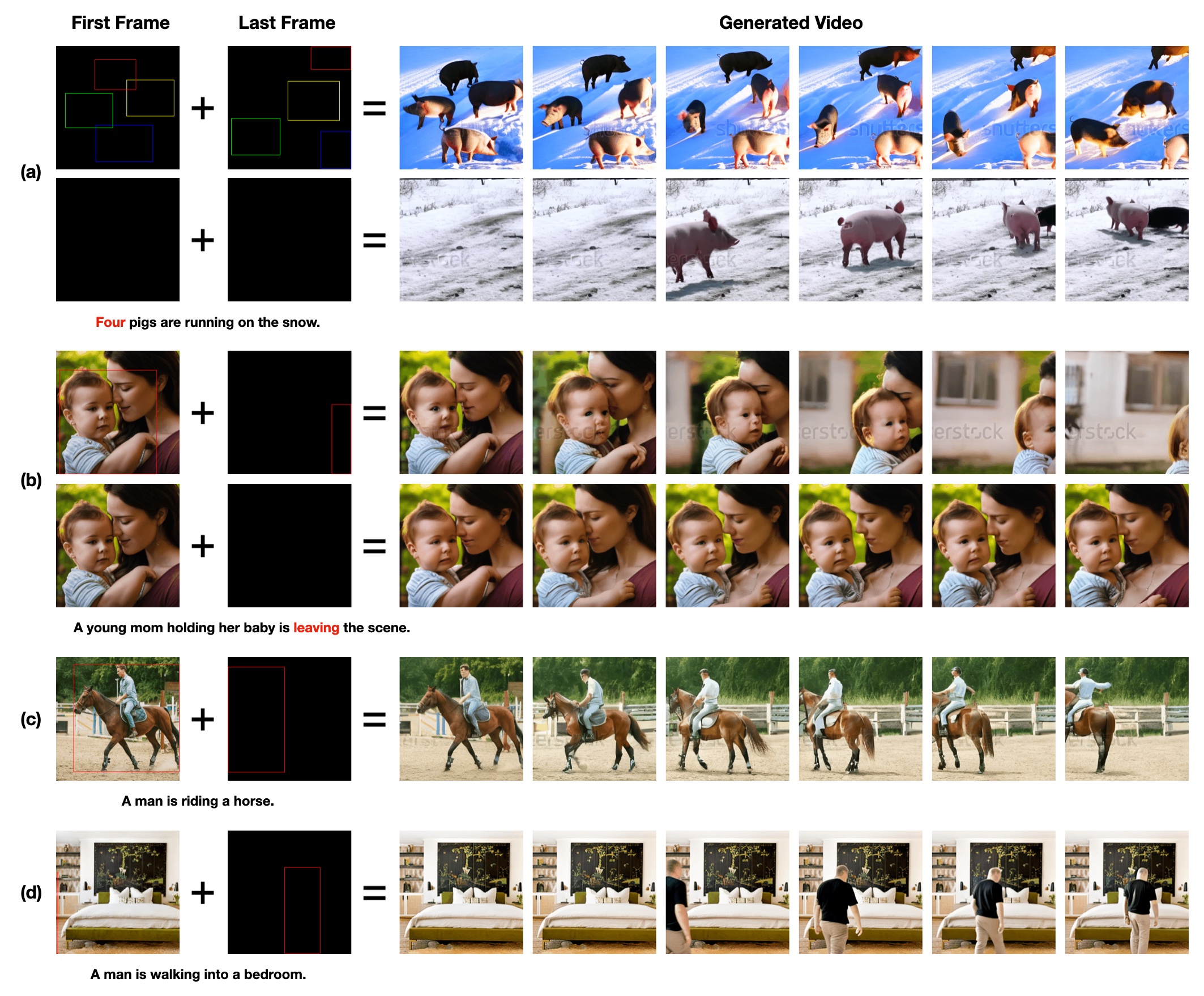}
\caption{Case study: (a) Generation and motion control based on four boxes; (b) A motion that affects significant portion of the frame; (c) Box defined on a combination of objects (e.g., ``a man on a horse''); (d) Adding new objects to the scene.}
\label{fig:case-study}
\vspace{-10pt}
\end{figure*}

\subsection{Human Evaluation}

We conducted a user preference study with four human raters on 100 samples. In each session, they were shown two videos in a random order: one generated by the base model (PixelDance), which uses a text prompt and the first frame as input, and the other by the Boximator model, which additionally uses box constraints. The raters were asked to evaluate their preference based on video quality and motion control. Detailed criteria for evaluation are presented in Appendix~\ref{sec:human-eval}. As indicated in Table~\ref{table:human-eval}, the Boximator model was preferred by a significant margin. It excelled in motion controls in 76\% of the cases, outperformed by the base model in only 2.2\% of the cases. The Boximator model's video quality was also favored (+18.4\%), likely due to the dynamic and vivid content resulting from box constraints. See Appendix~\ref{sec:human-eval} for some sample videos.

\subsection{Ablation Study}\label{sec:ablation}

We carry out ablation studies to understand the effect of our design choices. Initially, we exclude self-tracking from our training process. This means we train the model to predict the original video without any visible bounding boxes. We observe that omitting self-tracking greatly challenges the model's ability to associate control tokens with the corresponding objects. As shown in Table~\ref{table:ablation}, the average precision (AP) under box constraints falls drastically, reaching a level that is only slightly better than the AP without box constraints.

Next, we examine the role of using soft boxes during inference. According to the standard inference method described in Section~\ref{sec:inference}, we insert relaxed soft boxes in frames 2-15, where the user does not specify any box constraints. Table~\ref{table:ablation} indicates that removing these relaxed soft boxes (by replacing their control tokens with $\bm{t}_{\rm null}$) leads to a significant decrease in average precision scores. We hypothesize that the inserted soft boxes act as a rough guide for movement directions. Without this guide, the model tends to make more mistakes.

Finally, we examine the impact of freezing the base model weights. For comparison, we trained a new model in which all parameters of the U-Net were optimized. We find that the new model generates videos of roughly the same quality, resulting in similar FVD scores as the standard model in Table~\ref{table:msr-vtt-results}. When it comes to motion control precision, as shown in Table~\ref{table:ablation}, this new model scored similarly as the default one on MSR-VTT, and lower on ActivityNet. In summary, our results suggest that it's not necessary to train all the U-Net parameters.
\subsection{Case Study}

In this section, we highlight the model's capability of handling complex scenarios. Figure~\ref{fig:case-study}(a) demonstrates a generation task based on four boxes. Boximator successfully populates each box with the target object (a pig) as specified in the text prompt. This contrasts with the second row, where the model without box constraint only produces two pigs. Indeed, previous research has found that text-conditioned diffusion models struggle with precise object count control without box constraints~\cite{yang2023reco}.

Figure~\ref{fig:case-study}(b) illustrates a dynamic scene where a baby is moved across the entire frame. The box has guided the model to generate the motion, which appeared to be challenging to generate without box constraints (see the next row). Figure~\ref{fig:case-study}(c) highlights the generalizability of box-based visual grounding. Here, a user wants to control an object combination: a man on a horse. The model interprets this constraint, moving the composite object towards the frame's left edge. Finally, Figure~\ref{fig:case-study}(d) showcases the model's capability to introduce a new object into a scene. The user indicates the entry point of a man by drawing a segment line along the left border. The model successfully directs a man entering from the left edge, stopping at the center.

\section{Conclusion}

We proposed Boximator, a genral approach to controlling object motion in video synthesis. Boximator utilizes two types of boxes to allow users to select arbitrary objects and define their motions without entering extra text. It can be built on any video diffusion model without modifying the original model weights, thus its performance can improve with evolving base models. Additionally, we proposed a self-tracking method that significantly simplifies the training for the control module. We believe that our design choices and training techniques can be adapted to enable other forms of control, such as conditioning with human poses and key points.

\section*{Ethical and Social Risks}
Video generation technologies, especially advanced video diffusion models, carry potential ethical and social risks. These include the creation of deepfakes, which can lead to misinformation and privacy violations; biases in AI-generated content, potentially leading to unfair or discriminatory outcomes; and impacts on intellectual property and creative industries, possibly undermining the value of human creativity. It's crucial for developers and users of these technologies to be aware of these risks and ensure their responsible use.
\bibliography{refs}
\bibliographystyle{ieeenat_fullname}

\newpage
\appendix
\onecolumn
\section{More Implementation Details}\label{sec:appendix-implementation-details}

\paragraph{Control Module} We follow NeRF~\cite{mildenhall2021nerf} to use Fourier embeddings to encode box coordinates, object ID and the hard/soft flag. We make sure that all input dimensions are scaled between 0 and 1. For any given input $x$ within this range, its Fourier embedding is defined as:
\[
{\rm Fourier}(x) = [\cos(x\cdot 100^{0/8}),\dots, \cos(x\cdot 100^{7/8}), \sin(x\cdot 100^{0/8}), \dots, \sin(x\cdot 100^{7/8})].
\]
We combine these Fourier embeddings of each input to form the overall embedding, which has a dimension of 128. As mentioned in Section~\ref{sec:architecture}, these embeddings are then processed through a multi-layer perceptron (MLP). This MLP has three hidden layers, each with a dimension of 512. Finally, the output control token is adjusted to match the dimension of the visual token, which is 1024.

\paragraph{Training \& Inference Details} Our models train on 16-frame sequences with a resolution of 256x256 pixels, running at 4 frames per second. We limit the maximum number of objects to $N=8$. The training uses the Adam optimizer, with a batch size of 128 across 16 NVIDIA Tesla A100 GPUs. As outlined in Section~\ref{sec:training-procedure}, training occurs in three stages: 50k iterations for stage 1, 50k iterations for stage 2, and 10k iterations for stage 3. We use $2\times10^{-4}$ learning rate for the first stage, and $3\times10^{-5}$ for later stages. All stages use linear learning rate scheduler with 7,500 warm-up steps. Since box conditioning is optional, we use 25\% of our training data from videos without any box annotation. Since first frame conditioning is also optional, we let half of the training samples include the video's first frame as a condition.

For all experiments, we use the DDIM inference algorithm~\cite{song2020denoising} with 50 inference steps. To enable classifier-free guidance, we construct negative conditions by substituting every control token with $\bm{t}_{\rm null}$. We set the classifier-free guidance scale to be 9. 

\section{Results on UCF-101}\label{sec:appendix-ucf}

\begin{table*}[ht]
\centering
\begin{tabular}{l l | c c} 
\toprule
\textbf{Models} & Extra Input & FVD($\downarrow$) & mAP/AP$_{50}$/AP$_{75}$($\uparrow$) \\ \hline
MagicVideo~\cite{zhou2022magicvideo} & - & 699 & - \\
LVDM~\cite{he2022latent} & - & 641 & - \\
ModelScope~\cite{wang2023modelscope} & - & 410 & - \\
Make-A-Video~\cite{singer2022make} & - & 367 & - \\
VidRD~\cite{gu2023reuse} & - & 363 & - \\
PYOCO~\cite{ge2023preserve} & - & 355 & - \\
Dysen-VDM~\cite{fei2023empowering} & - & 325 & - \\
PixelDance~\cite{zeng2023make} & - & {\bf 242} & - \\
Stable Video Diffusion~\cite{blattmann2023stable} & - & {\bf 242} & - \\ \hline
\hline
\multirow{4}{*}{PixelDance + Boximator} & - & 270 & 0.060/0.127/0.044 \\
                  & Box & 263 & 0.228/0.354/0.229 \\
                  & F0 & 132 & 0.171/0.272/0.163 \\
                  & F0 + Box & 142 & 0.284/0.419/0.279 \\ \hline
\multirow{4}{*}{ModelScope + Boximator} & - & 310 & 0.063/0.131/0.047\\
                  & Box & 311 & 0.192/0.308/0.184 \\
                  & F0 & 196 & 0.132/0.223/0.119 \\
                  & F0 + Box & 194 & 0.212/0.343/0.205 \\ \bottomrule
\end{tabular}
\caption{Zero-shot results on UCF-101.}
\label{table:ucf-results}
\end{table*}

We follow the experiment settings of PixelDance~\cite{zeng2023make} to evaluate on UCF-101. Specifically, we sampled 2,048 videos from the UCF-101 test set, generating descriptive text prompts for each of them, and then generated 10,240 16-frame videos. We compute the FVD real features from the original 2,048 videos by sampling 16 frames from each video. Reference bounding boxes were automatically annotated using the same method as for MSR-VTT. Given generated videos, we employed DINO+DEVA for bounding box detection and computed average precision (AP) scores. It's noteworthy that UCF-101's prompts are more detailed than those for MSR-VTT and ActivityNet. Since the automatic annotation uses the text prompt to extract object names, the longer prompts lead to more, albeit noisier, boxes per video.

Table~\ref{sec:appendix-ucf} presents our UCF-101 results, showing trends consistent with MSR-VTT. The Boximator model roughly maintained or improves the FVD scores compared to the base model. While using the first frame (F0) as a condition notably boosted FVD scores, box constraints had minimal impact to FVD, likely due to the noisier nature of UCF-101's boxes.

In all scenarios, using box constraints significantly increased AP scores, echoing results from MSR-VTT and ActivityNet. However, the absolute AP values on UCF-101 were lower than on the other datasets, probably due to the lower quality of box annotations.

\section{Human Evaluation Details}\label{sec:human-eval}

We selected 100 high-quality videos featuring prominent camera or object movements from WebVid (excluded from training set) and manually annotated their bounding boxes. Then we generate new videos using both the standard PixelDance model and PixelDance+Boximator, with the video caption and the first frame taken as inputs. The Boximator model additionally used bounding boxes from the first and last frames. Four human raters assessed the regenerated videos, marked as ``Video 1'' and ``Video 2,'' presented in a randomized order to obscure the generating model. Raters evaluated the videos for quality and motion control, choosing between ``Video 1 is better,'' ``Video 2 is better,'' or ``no preference.''

\paragraph{Video Quality} Raters evaluated each video for visual distortions, blurs, or other quality defects, and for temporal inconsistencies, such as inconsistent object appearances across frames. In cases where both videos were free from these issues, raters favored the video with richer content. For instance, when comparing two videos where one exhibits interesting motion and the other remains mostly stationary, raters are expected to favor the more dynamic one.

\paragraph{Motion Control} The evaluation focused on whether each video satisfied motion constraints set by the bounding boxes in the initial and final frames. Preference was given to the video meeting these constraints. If both or neither video met the constraints, raters are expected to select ``no preference.''

\begin{figure*}[ht]
\centering
\includegraphics[width=\textwidth]{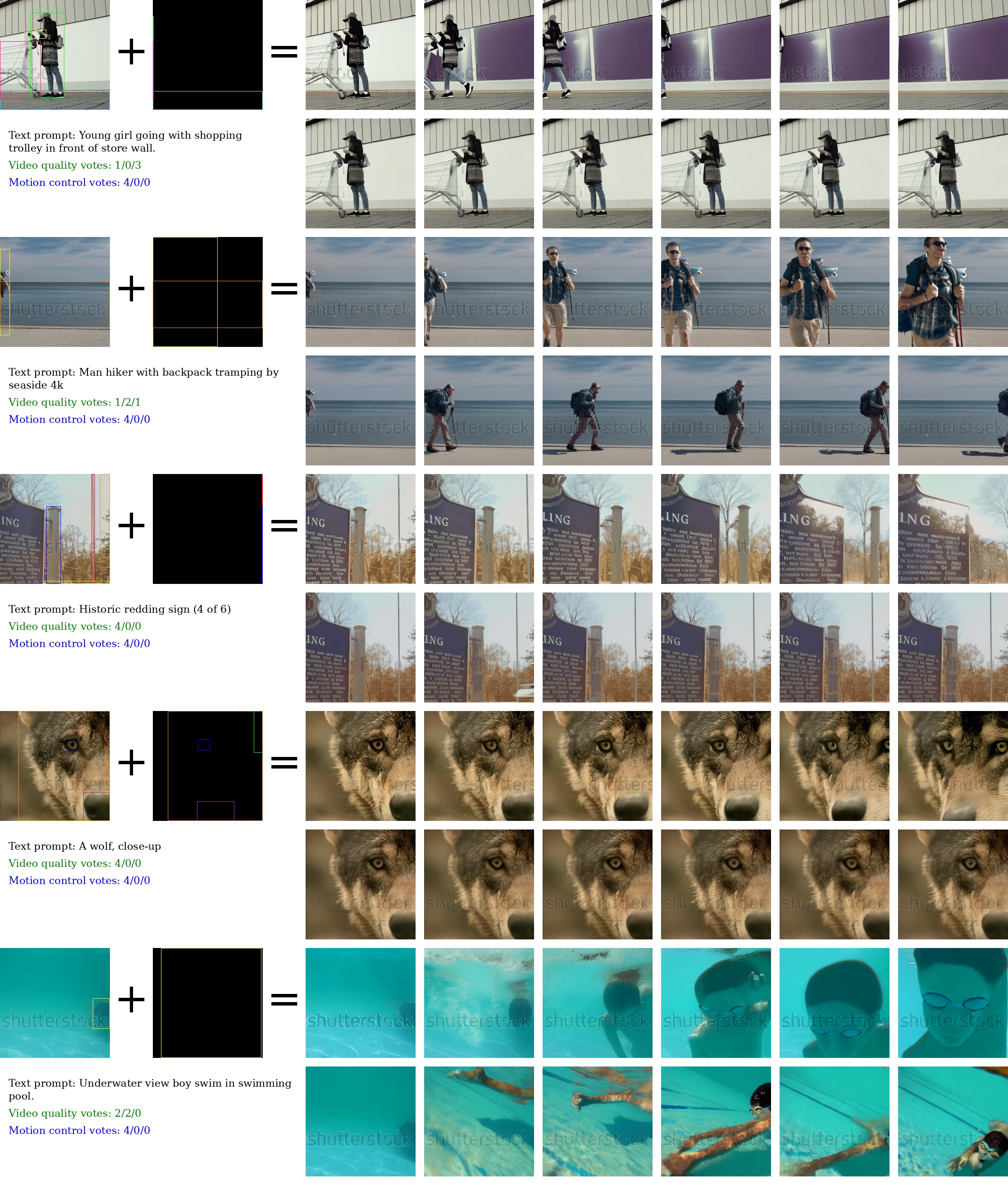}
\caption{Sample videos from human evaluation (Part 1). Each group displays two rows: the first generated by the Boximator model and the second by the base model. Vote results are denoted as X/Y/Z, indicating raters' preferences: X for Boximator model, Y for no preference, and Z for base model.}
\label{fig:video_samples-1}
\end{figure*}

\begin{figure*}[ht]
\centering
\includegraphics[width=\textwidth]{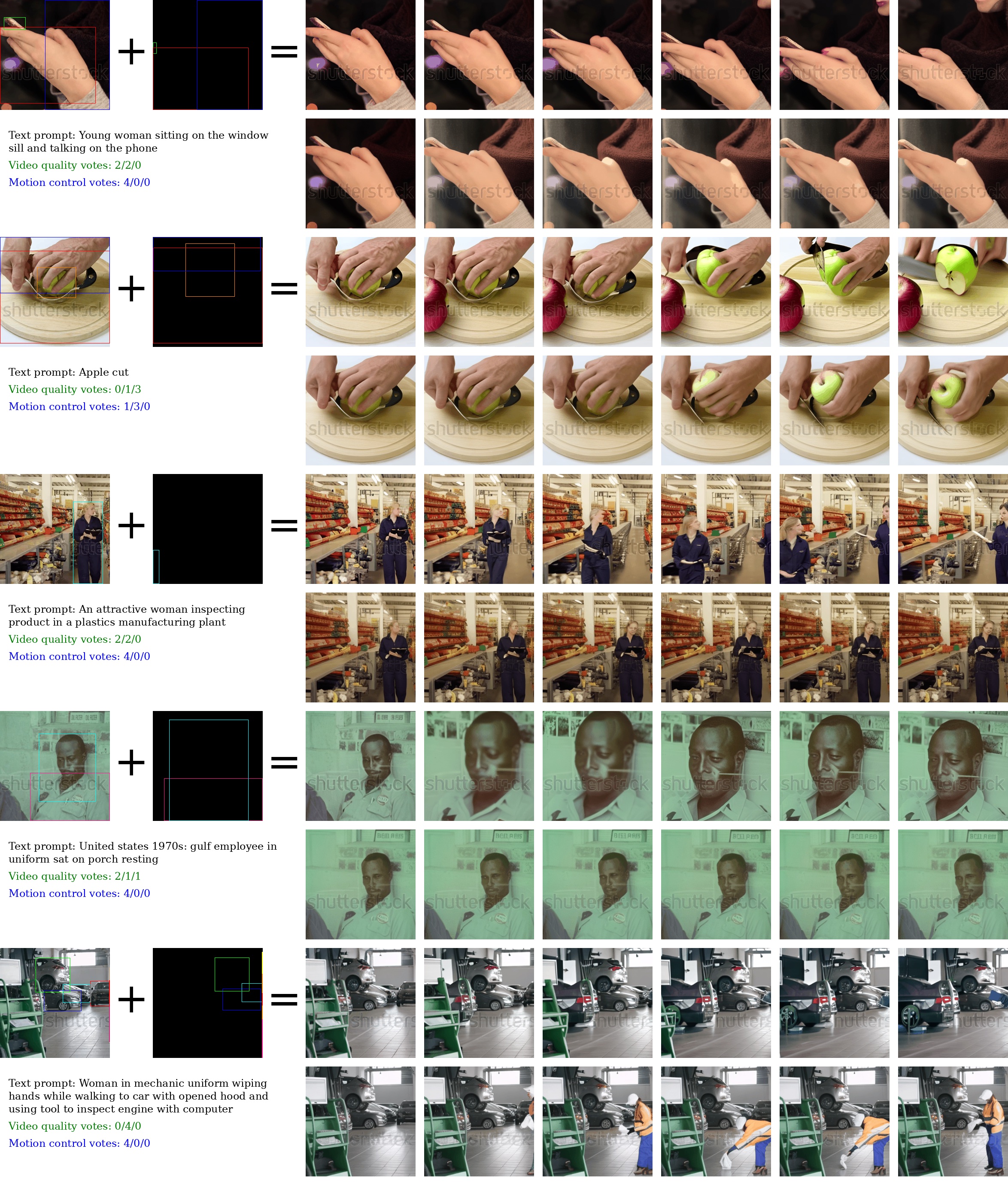}
\caption{Sample videos from human evaluation (Part 2).}
\label{fig:video_samples-2}
\end{figure*}

\begin{figure*}[ht]
\centering
\includegraphics[width=\textwidth]{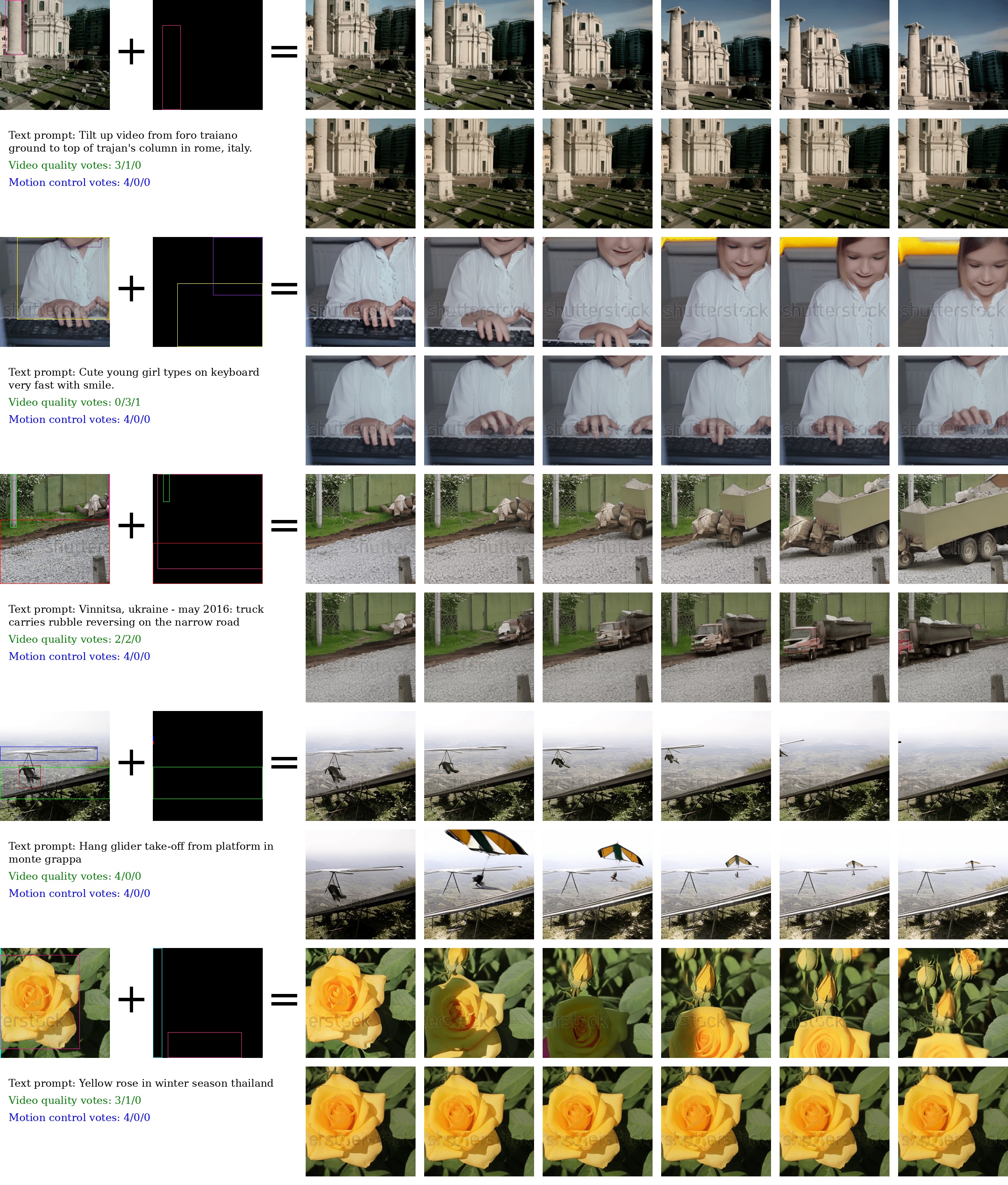}
\caption{Sample videos from human evaluation (Part 3).}
\label{fig:video_samples-3}
\end{figure*}

Some sample videos and their evaluations results are displayed in Figures~\ref{fig:video_samples-1} to \ref{fig:video_samples-3}.

\end{document}